# **GridForge:** An Open Physics-Informed Digital Twin and Benchmark Suite for UK 11 kV Underground Distribution Cables

Lilliane Linnet Musoke
Independent Researcher
lylliamusoke@gmail.com

**Abstract**

The UK distribution-cable monitoring sector is data-rich but framework-poor: each Distribution Network Operator (DNO) collects megabytes of telemetry per asset per day, but there is no public benchmark dataset, no shared reference model, and no commonly agreed evaluation protocol for benchmarking sectoral, data-intensive models for operational decision support; notably for maintenance planning and prioritisation. We address this gap with **GridForge**, an open-source physics-informed digital twin and benchmark suite for fault diagnosis and prediction, integrating four building blocks: (i) a verified steady-state and transient thermal model based on IEC 60287 and IEC 60853; (ii) a calibrated synthetic dataset of 64 cables (covering 320 cable-years) representing four canonical UK 11 kV thermal archetypes and four condition modes; (iii) a Physics-Informed Neural Network (PINN) surrogate that satisfies the algebraic balance defined by IEC 60287 during training and reaches an RMSE of 0.148 °C on an independent held-out set against the reference analytical oracle; and (iv) a five-task benchmark suite that includes three reference baseline models (pure physics, gradient-boosted trees, and the PINN). All numerical parameters are sourced from public standards, the code is licensed under the MIT licence, and the trained PINN converges in only two minutes (as run on the computing platform in Section 10.1), enabling independent verification by DNO engineering and research practitioners. At present only the virtual-sensor task (T4) discriminates between the reference baselines; the other tasks tie or are undefined on this test split, so extending the baselines with task-specific logic is a priority future enhancement. For open validation, the dataset, test sets and models are made available at https://github.com/mlily2024/Gridforge

## 1 Introduction

About 130,000 km of 11 kV underground distribution cable is operated by the 14 Distribution Network Operator (DNO) licence areas in the United Kingdom (Energy Networks Association 2024). Replacing a medium-voltage cable section is a six-figure intervention: on the published connection unit costs of the network operator, a 500 m 11 kV section is costed at approximately £150,000, rising above £200,000 for longer sections (National Grid Electricity Distribution 2023). Reactive replacement after an in-service failure, which requires emergency excavation, is markedly more expensive than the same work planned within a maintenance window. There is therefore an economic case for predictive, condition-based cable replacement, reinforced by the RIIO-ED2 framework of the UK regulator Ofgem, which links Customer Interruption (CI) and Customer Minutes Lost (CML) penalties to unplanned outages (Ofgem, 2022).

The technical case for condition-based cable replacement is more challenging. Over the past two decades, an asset manager would typically spend around £2 million on partial-discharge monitoring (International Electrotechnical Commission, 2000), distributed temperature sensing, and SCADA integration, and yet there is no decision-support tool to help plan and prioritise maintenance across the entire cable network end-to-end. There is no answer to the question that the network manager is entitled to ask, *"given my £2 million capital budget, which 12 cable sections should I replace this year, and what failure risk does* *deferring the rest carry?"* The reasons for the absence of a maintenance-planning support tool are systemic:

1. **No public benchmark dataset.** DNOs cannot release operational monitoring data due to commercial sensitivity, cybersecurity constraints, and the geographic specificity of fault records. Without a shared dataset, every published technique is evaluated on its author's private data and cannot be reproduced or compared.
2. **No shared reference model.** The closed-form rating equations of IEC 60287 are widely known but only sporadically implemented as open code. Open implementations of the transient lumped model (IEC 60853) are even rarer.
3. **No agreed-upon evaluation protocol.** Failure prediction, remaining useful life regression, anomaly detection, and counterfactual attribution each have well-developed evaluation traditions in adjacent fields (cardiology, structural health monitoring, econometrics). The distribution-cable literature has not yet adopted any of them at scale.

GridForge addresses these challenges by providing an integrated, full-stack open-source decision-support tool for cable network maintenance planning. It offers a reproducible benchmarking capability for cable-segment failure prediction and, hence, maintenance decision support that can be deployed today by any DNO or commercial operator, while the sector addresses the longer-term economic sustainability challenges of the distribution network. Reproducible, physics-grounded digital twins are increasingly regarded as core infrastructure for future power-system operation (Zomerdijk et al., 2024).

Our contributions are:

1. **An open physics oracle: verified** open-source implementations of the IEC 60287 steady-state and IEC 60853-2 transient thermal models, a coaxial electric-field solution, and Crine dielectric-ageing kinetics (Section 3).
2. **A calibrated synthetic benchmark dataset: 64 cables** (320 cable-years) of hourly telemetry with sealed ground-truth labels, every numerical parameter cited to a public standard (Section 4).
3. **A physics-informed surrogate: a compact PINN** trained against the oracle, with published validation (0.148 °C RMSE on an independent held-out set) and full reproducibility (Section 5).
4. **A sealed five-task benchmark: heterogeneous** prediction tasks with three reference baselines scored under a fixed evaluation protocol (Section 6).

## 2 Related works

International grid-asset decision platforms such as Copperleaf C55, DNV Synergi Grid, IBM Maximo APM, and Siemens Spectrum Power Asset Health Manager operate at the asset-portfolio level, integrating with utilities' Condition Monitoring Management Systems (CMM) and Geographic Information Systems (GIS) as an enterprise SaaS (Mirhosseini and Keynia, 2021). None has released a public benchmark or a reference physics model.

UK-specific monitoring vendors, such as Kestrel, Camlin, EA Technology, Lucy GridKey and Synaptec, focus on sensing and per-asset alerting; their analysis is predominantly threshold-based, and machine-learning methods appear mainly in research rather than in shipped products (CIGRE Working Group B1.42 2019; Kumar et al., 2024).

Research on physics-informed neural networks (Raissi et al., 2019; Cuomo et al. 2022) for power systems (e.g. (Misyris et al. 2020)) has so far focused on power-flow and transient-stability problems rather than cable-asset health. The most widely adopted ageing model for cables is the Crine unified model (Crine 2005), which is rarely incorporated in machine-learning pipelines (Mazzanti & Marzinotto, 2013). Recent surveys map the rapid growth of physics-informed neural networks across power-system applications (Huang & Wang, 2023), and dedicated studies of PINNs as grid surrogates report strong physical-constraint satisfaction alongside some accuracy degradation in extreme operating regimes (Cestero et al. 2025).

To our knowledge, GridForge is the first open release of:

- a calibrated synthetic dataset for UK 11 kV underground cables;
- a benchmark suite with sealed test labels;
- a physics-informed neural network with published validation against the IEC 60287 analytical oracle.

While simulation-derived benchmarks are well established in adjacent prognostics domains, notably the NASA C-MAPSS turbofan run-to-failure dataset (Saxena et al., 2008), no equivalent open benchmark exists for UK 11 kV distribution cables.

## 3 Methodology overview and forward model

GridForge is built and evaluated as a four-stage pipeline. **(1) Forward model (physics oracle):** we implement and verify a first-principles cable model, comprising the IEC 60287 steady-state and IEC 60853-2 transient thermal solutions, a coaxial electric-field solution, and Crine dielectric-ageing kinetics; this serves as the trusted analytical reference (this section). **(2) Dataset generation:** we drive the oracle to synthesise calibrated cable-year telemetry across the cable archetypes, condition modes, and representative UK load and weather profiles, with deterministic splits and sealed test labels (Section 4). **(3) Surrogate training:** we train a physics-informed neural network against the oracle to obtain a fast, differentiable conductor-temperature predictor (Section 5). **(4) Benchmark evaluation:** we define five sealed-label tasks with three reference baselines and score them under a fixed protocol (Section 6). Every stage is reproducible from documented constants and fixed random seeds (Section 10).

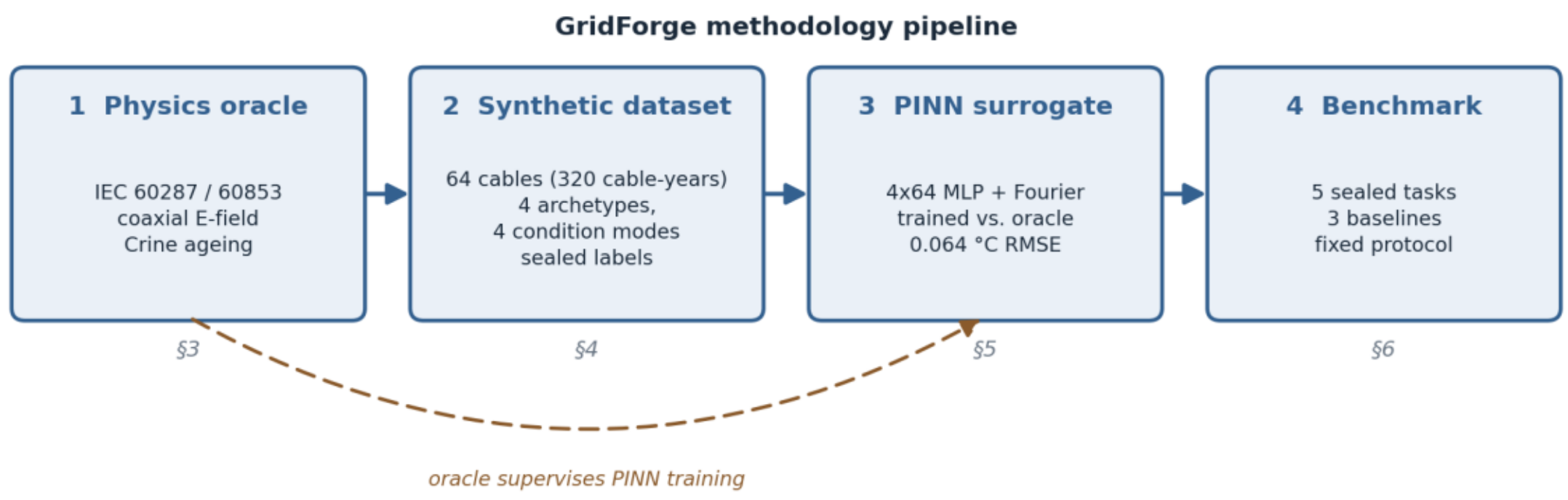

***Figure 1.*** *The GridForge methodology pipeline. A verified physics oracle (Section 3) generates a calibrated synthetic dataset (Section 4) and supervises the training of a physics-informed neural-network surrogate (Section 5); a five-task benchmark then scores the surrogate against reference baselines under a fixed protocol (Section 6).*

GridForge models a buried 11 kV underground cable as a coupled thermal-electrical-degradation system. The three governing equations are summarised below; full derivations follow IEC 60287-1-1 (International Electrotechnical Commission 2006) and IEC 60287-2-1 (International Electrotechnical Commission 2015). The IEC 60287 rating method formalises the classical Neher-McGrath thermal-resistance analysis (Neher

& McGrath, 1957); Anders (1997) provides a comprehensive treatment of steady-state and transient ampacity computations.

### 3.1 Steady-state thermal model

The IEC 60287-1-1 conductor temperature rise above ambient is

$$\Delta\theta_c = (I^2R + 1/2\,W_d)T_1 + (I^2R + W_d)\,nT_2 + (I^2R + W_d)\,(1+\lambda_1)\,nT_3 + (I^2R + W_d)\,(1+\lambda_1 + \lambda_2)\,nT_4,$$

where $I$ is the per-conductor current, $R = R(\theta_c)$ the a.c. conductor resistance, $W_d$ the per-phase dielectric loss, $n$ the number of load-carrying conductors, $\lambda_1$ and $\lambda_2$ the sheath and armour loss factors, and $T_1, T_2, T_3, T_4$ the per-unit-length thermal resistances of the insulation, separator beds, jacket, and surrounding medium. Closed-form expressions for $T_1, T_3, T_4$ in terms of cable geometry and burial conditions are given in (International Electrotechnical Commission 2015); for a single buried cable

$$T_4 = \frac{\rho_s}{2\pi}\ln\left(2u + \sqrt{4u^2 - 1}\right), \qquad u = \frac{2L_b}{D_e},$$

where $\rho_s$ is the soil thermal resistivity, $L_b$ the burial depth and $D_e$ the overall outside diameter of the cable. The system is implicit because $R$ depends on $\theta_c$; we solve by fixed-point iteration, which converges in 5–10 steps for typical operating conditions.

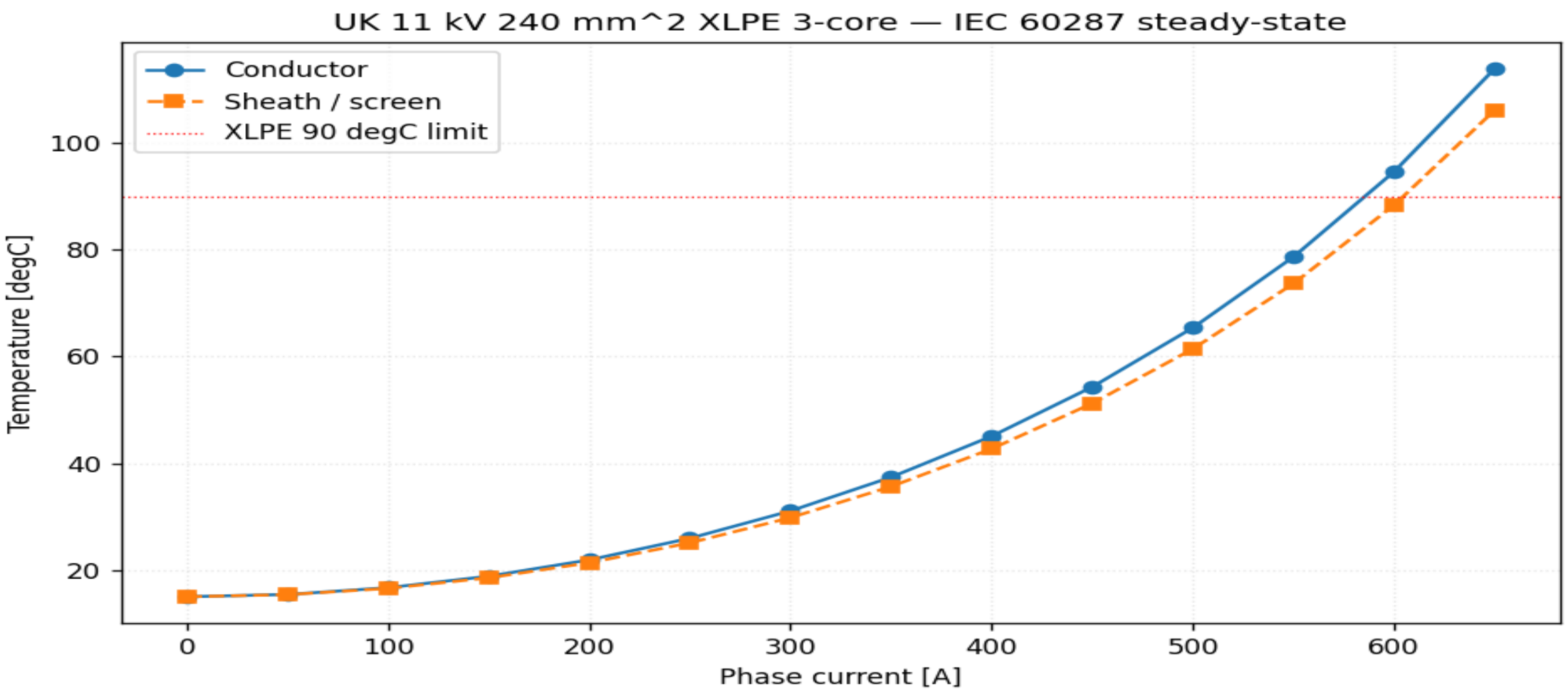

*Figure 2. Steady-state conductor temperature predicted by the IEC 60287 fixed-point solver for the UK 11 kV 240 mm² Cu XLPE 3-core archetype at 0.8 m burial depth and 1.0 K·m/W soil thermal resistivity. The super-linear rise above 350 A reflects the temperature dependence of the conductor a.c. resistance $R(\theta_c)$.*

### 3.2 Transient thermal model (lumped first-order)

For load-following studies on hour-to-day timescales, we use the IEC 60853-2 simplified single-node lumped form (International Electrotechnical Commission 2008):

$$\frac{C_c}{n}\frac{d\theta_c}{dt} = (I^2R(\theta_c) + W_d) - \frac{\theta_c - \theta_a}{R_{\text{total}}},$$

with $R_{\text{total}} = T_1 + n(1+\lambda_1)(T_3+T_4)$ and $C_c$ the per-unit-length thermal capacitance of the cable cross-section. The first-order time constant $\tau = R_{\text{total}} C_c/n$ sits around 64 minutes for a UK 11 kV 240 mm² Cu XLPE 3-core cable buried at 0.8 m. Steady-state recovery of the Section 3.1 balance at $d\theta_c/dt = 0$ is exact by construction. Transient thermal-circuit modelling of cables remains an active research area, including recent formulations that account for axial heat dissipation (Qin et al., 2025).

### 3.3 Electric field

For a coaxial XLPE-insulated cable with conductor outer radius $r_c$ and screen inner radius $r_s$, the radial field under a.c. operation is the Laplace solution

$$E(r) = \frac{U_0}{r\ln(r_s/r_c)},$$

with maximum at the conductor surface ($E_{\max} = E(r_c)$). The Crine ageing model is driven by $E_{\max}$.

3.4 Crine ageing kinetics

Crine's 2005 unified model (Crine, 2005) gives mean time-to-failure under constant electric-field and thermal stress as

$$L(E,T) = L_{\text{ref}}\left(\frac{E_{\text{ref}}}{E}\right)^n \exp\left[\frac{\Phi}{k_B}\left(\frac{1}{T} - \frac{1}{T_{\text{ref}}}\right)\right],$$

parameterised by an inverse-power-law exponent $n$, an Arrhenius activation energy $\Phi$, and a calibration triple $(E_{\text{ref}}, T_{\text{ref}}, L_{\text{ref}})$. We default to the XLPE design point $(4\ \text{MV/m}, 90\ °\text{C}, 40\ \text{years})$. Under time-varying stress, cumulative damage follows a continuous generalisation of Miner's linear cumulative-damage rule (Miner 1945)

$$D(t) = \int_0^t \frac{d\tau}{L(E(\tau), T(\tau))},$$

with failure at $D = 1$. The remaining useful life under an assumed forward stress $(E_f, T_f)$ is $\text{RUL} = (1 - D_{\text{now}})\, L(E_f, T_f)$.

## 4 The synthetic dataset

Training and evaluating on a calibrated synthetic dataset, rather than field measurements, is established practice in physics-informed surrogate modelling (where networks are supervised by the governing equations and their analytical solutions (Raissi et al. 2019; Misyris et al. 2020)) and in prognostics, where the most widely used benchmarks are simulation-derived, notably the NASA C-MAPSS turbofan run-to-failure dataset (Saxena et al. 2008). Two properties motivate it here: the objective of the surrogate is to reproduce a trusted analytical standard, so validation against that standard is the appropriate criterion; and real cable-failure data is scarce and proprietary, whereas a synthetic dataset specified entirely by published constants and deterministic seeds is reproducible by any third party.

GridForge ships a 64-cable benchmark dataset (Gridforge-mini) constructed from a 6-tuple specification:

- **archetype** ∈ {240 mm² XLPE 3c, 95 mm² XLPE 3c, 300 mm² XLPE 1c, 240 mm² PILC 3c}
- **load profile** ∈ {residential, commercial, industrial, mixed}
- **condition mode** ∈ {healthy, water_ingress, thermal_ageing, accelerated_dielectric}

- **weather seed** (UK Met Office 1991–2020 soil-temperature climatology with stochastic noise)
- **duration** (default 5 years, hourly resolution)
- **deterministic random seed**

Every numerical parameter (cable geometry, material thermal resistivities, soil-temperature climatology, Crine ageing constants) is cited to a public source: BS EN 60228 for conductor sizes (British Standards Institution 2005), BS 7870 for UK distribution-cable practice (British Standards Institution 2011), IEC 60287-2-1 for thermal resistivities (International Electrotechnical Commission 2015), Met Office UK 1991–2020 soil climatology (UK Met Office 2024) and IEEE Std 1407-2007 for accelerated-ageing parameters (IEEE, 2007). No employer data, no NDA-protected data, and no commercially confidential parameters are included in the dataset.

Each cable-year is generated by the following procedure. (1) Sample a specification 6-tuple. (2) Build the physical parameters of the archetype from the cited public sources. (3) Generate an hourly load-current series from the load profile and an hourly ambient and soil-temperature series from the weather climatology with added stochastic noise. (4) Integrate the coupled thermal and Crine-ageing equations of Section 3 with the fixed-point solver to obtain the conductor-temperature and cumulative-damage trajectories. (5) Apply the condition-mode injector: *healthy* leaves the trajectory unchanged, while *water_ingress*, *thermal_ageing*, and *accelerated_dielectric* accelerate degradation through their characteristic mechanism. (6) Write the resulting hourly telemetry and, separately, the sealed ground-truth labels (failure time, damage trajectory, and driver attribution).

The deterministic SHA-256-based train/val/test split of the dataset (70/15/15) keeps each cable atomic across its telemetry stream. Sealed test labels live in a separate ground_truth/failure_times.csv file so a benchmark submission can withhold them at evaluation time.

A minimal Gridforge-mini build with 64 cables × 5 years × hourly resolution comprises approximately 2.8 million samples and around 200 MB of uncompressed CSV; production-scale deposits of ~1000 cable-years (planned as future work) fit within the ~10 GB Zenodo deposit limit.

## 5 Physics-informed neural network surrogate

The IEC 60287 fixed-point solver of Section 3.1 is exact but slow (millisecond per call); for decision-engine queries that evaluate $\theta_c$ thousands of times per fleet ranking, a fast differentiable surrogate is worthwhile. We train an MLP with sinusoidal positional encoding (Tancik et al., 2020):

- **inputs** $(I, \theta_a, \rho_s) \in \mathbb{R}^3$
- **output** $\theta_c \in \mathbb{R}$
- **architecture** four hidden layers of 64 units, tanh activations
- **encoding** Fourier features at $n_{\text{freqs}} = 4$ octaves
- **parameterisation** $\theta_c = \theta_a + \text{softplus}(\text{MLP}(\text{enc}(x)))$, enforcing $\theta_c \geq \theta_a$ by construction

Total parameters: 14 337.

### 5.1 Loss

We train against IEC 60287 oracle data with a combined loss

$$\mathcal{L} = w_d\,\mathcal{L}_{\text{data}} + w_p\,\mathcal{L}_{\text{phys}}$$

where $\mathcal{L}_{\text{data}}$is the mean-squared error against the oracle and $\mathcal{L}_{\text{phys}}$ is the squared residual of the IEC 60287 algebraic balance evaluated at the prediction of the network. Gradients flow through the prediction into the residual, so back-propagation enforces physics consistency directly.

The loss weights are adapted via gradient-norm balancing (Wang et al., 2022): every $K$ steps, $w_p \leftarrow \alpha w_p + (1-\alpha)\ \| \nabla\mathcal{L}_{\text{data}} \| / \| \nabla\mathcal{L}_{\text{phys}} \|$, keeping both terms on the same gradient scale throughout training. We use $\alpha = 0.9$, $K = 50$, and an initial $w_p = 10^{-3}$.

### 5.2 Training and validation

We sample 4000 training and 800 validation pairs uniformly from $I \in [50{,}600]$ A, $\theta_a \in [0{,}25]$ °C, $\rho_s \in [0.7{,}2.0]$ K·m/W, paired with the IEC oracle value for $\theta_c$. Adam at learning rate $2 \times 10^{-3}$, batch size 256, 1500 epochs trains in approximately two minutes, as run on the computing platform in Section 10.1. All architecture and optimisation hyperparameters were fixed a priori, with the loss-weight schedule following (Wang et al., 2022); no hyperparameter search was performed, since the surrogate already meets its accuracy target with a wide margin, as reported below. We report the **independent held-out RMSE of 0.148 °C** (1000 pairs drawn separately from the training and validation data) as the headline figure, because it measures generalisation on data used neither for training nor for checkpoint selection. For completeness, the best validation checkpoint reached 0.064 °C and the final-epoch validation RMSE was 0.131 °C; all three are comfortably inside the 0.5 °C acceptance target set a priori.

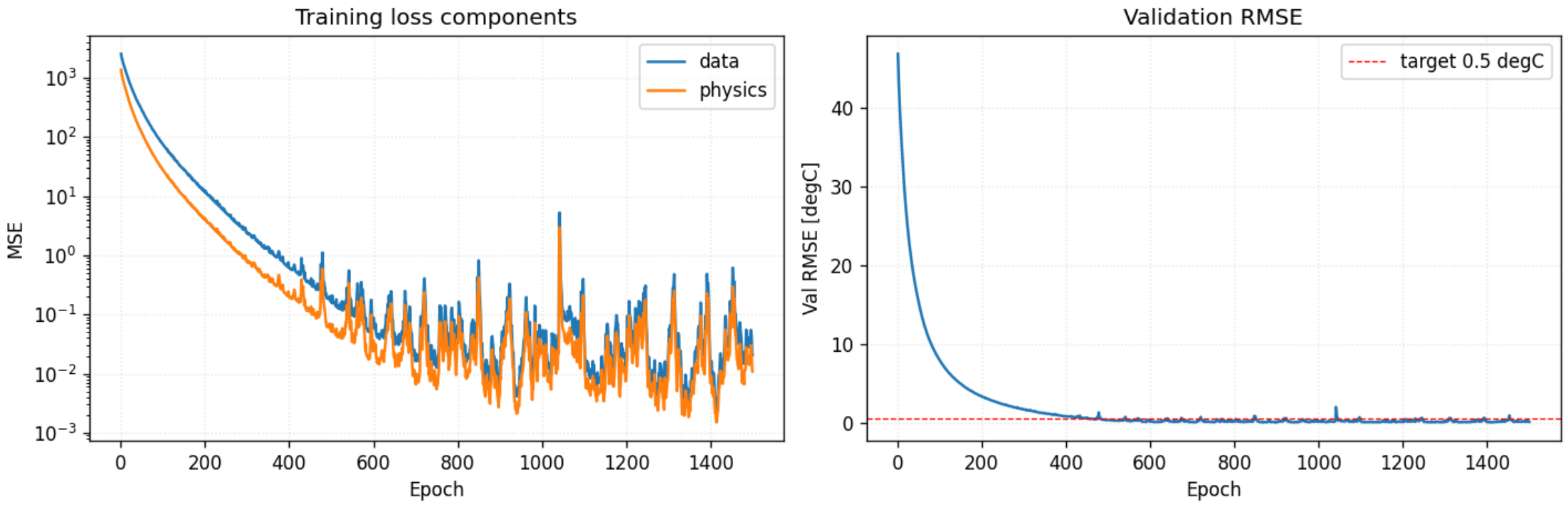

***Figure 3.*** *PINN training history over 1500 epochs. Top: data loss (MSE against the IEC 60287 oracle) and physics loss (squared IEC algebraic-balance residual evaluated at the prediction of the network), both on a log scale. Bottom: RMSE on the 800-pair validation set (the set used for checkpoint selection; distinct from the independent 1000-pair held-out set of Figure 4). The 0.5 °C acceptance target is reached by epoch 500 and the trajectory stabilises by epoch 1000.*

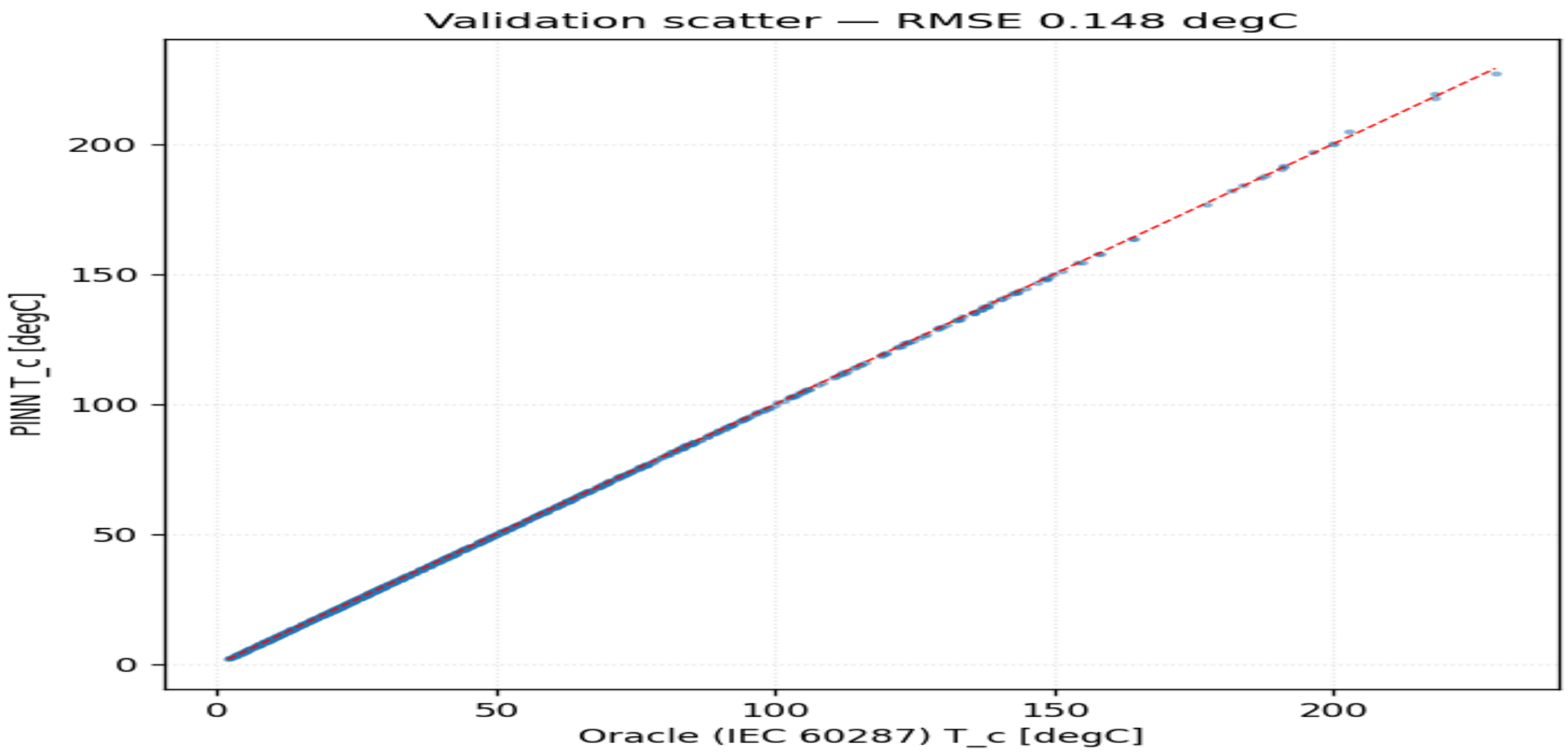

*__Figure 4.__ PINN predictions versus IEC 60287 oracle ground truth on an independent held-out set (n = 1000). All points lie close to the diagonal across the full input range $I \in [50,600]$ A, indicating no systematic bias. Held-out RMSE 0.148 °C.*

## 6 Benchmark suite

GridForge ships five sealed-test tasks (***Table 1***) and three reference baselines (***Table 2***). Tasks are deliberately heterogeneous: failure prediction is binary classification, RUL is regression, anomaly is per-step density estimation, virtual sensor is dense regression, and counterfactual is constrained-fraction regression. As a result, no single model architecture dominates by construction. Rigorous evaluation methodology for remaining-useful-life prediction is itself an active research topic (Wang et al., 2024).

Evaluation follows a single fixed protocol for every baseline and external submission: at test time, a model receives only the input view for a task and returns predictions, which are scored against sealed ground-truth labels withheld until scoring, and each task is summarised by the single headline metric in ***Table 1***. The headline metrics are those established for each task family: the Brier score (Brier, 1950) for probabilistic failure prediction (T1), mean absolute error for the time-to-threshold and counterfactual tasks (T2, T5), both root-mean-square error and mean absolute error for the conductor-temperature task (T4) reported together because they diverge under the heavy-tailed transient errors of that task, and precision at a fixed recall (Davis & Goadrich, 2006) for anomaly detection (T3).

**Table 1: Benchmark tasks.**

| ID | Task | Headline metric | Better |
|---|---|---|---|
| T1 | 60-day damage-threshold-crossing prediction from a 30-day window | Brier | Lower |
| T2 | Time-to-damage-threshold regression | MAE [years] | Lower |
| T3 | Per-day anomaly detection | Precision @ recall = 0.9 | Higher |

| ID | Task | Headline metric | Better |
|---|---|---|---|
| T4 | Conductor-temperature prediction | RMSE, MAE [°C] | Lower |
| T5 | Counterfactual driver attribution | MAE on fractions | Lower |

**Table 2: Reference baselines.**

| Baseline | Mechanism |
|---|---|
| IECOracleBaseline | Pure physics (IEC 60287 + Crine), no ML |
| GradientBoostedBaseline | Histogram-based gradient-boosted regression trees (Friedman, 2001) (scikit-learn HistGradientBoosting) on the three physical inputs: current, ambient temperature, and soil thermal resistivity |
| PINNBaseline | The Section 5 physics-informed neural network |

All three implement the same Baseline.predict(view, task) interface; external submissions subclass Baseline and are evaluated against the same sealed labels.

## 7 Reference results and discussion

Headline-metric scores on the gridforge-mini test split (11 cables, 482 130 telemetry samples, deterministic SHA-256 selection) are given for each baseline in ***Table 3***.

**Table 3: Leaderboard, headline metrics.**

| Task | IEC Oracle | Gradient-Boosted | GridForge PINN |
|---|---|---|---|
| T1 Brier | 0.228 | 0.228 | 0.228 |
| T2 MAE [yr] | – (no test cables crossed threshold) | – | – |
| T3 P@R=0.9 | 0.320 | 0.320 | 0.320 |
| T4 RMSE [°C] | 0.001 † | 10.39 | 10.54 |
| T4 MAE [°C] | 0.001 † | 8.77 | **5.84** |
| T5 MAE | 0.141 | 0.141 | 0.141 |

†The IEC Oracle T4 baseline returns the lumped-transient ground truth of the dataset and therefore serves as an upper-bound reference rather than a fair steady-state physics comparator (see Section 9).

Task T4 warrants a full reading. On root-mean-square error, the two learned surrogates are close and both far behind the IEC oracle reference (0.001 °C, itself an upper bound; see the note to Table 3): the gradient-boosted baseline scores 10.39 °C and the PINN 10.54 °C, so on RMSE the PINN is marginally behind. The distinction appears on mean absolute error, where the PINN improves on the gradient-boosted baseline by **33 %** (5.84 vs 8.77 °C) on identical inputs, both models receiving only current, ambient temperature, and soil thermal resistivity. The gap between RMSE and MAE for both surrogates is the signature of a heavy-tailed error distribution: neither steady-state surrogate reproduces the transient thermal lag carried in the T4 ground truth (Section 9, limitation 2), and those lag-driven outliers dominate the squared error that RMSE reports. MAE, which is robust to that tail, shows that the physics-informed prior yields a markedly better central-tendency fit on the same inputs. We therefore read T4 as evidence for physics-informed learning on the steady-state relationship, while being explicit that neither surrogate yet captures the transient component.

Tasks T1, T3, and T5 show no differentiation across baselines. This reflects the current reference implementations rather than the tasks themselves: for these three tasks the gradient-boosted and PINN baselines both defer to the shared downstream decision logic, so they return identical predictions. Task-specific implementations that would let the baselines diverge are planned as future work (Section 9). We report the tie rather than omit these tasks, so that the leaderboard reflects the present state of the reference baselines honestly.

## 8 Case studies

### 8.1 Diurnal load response

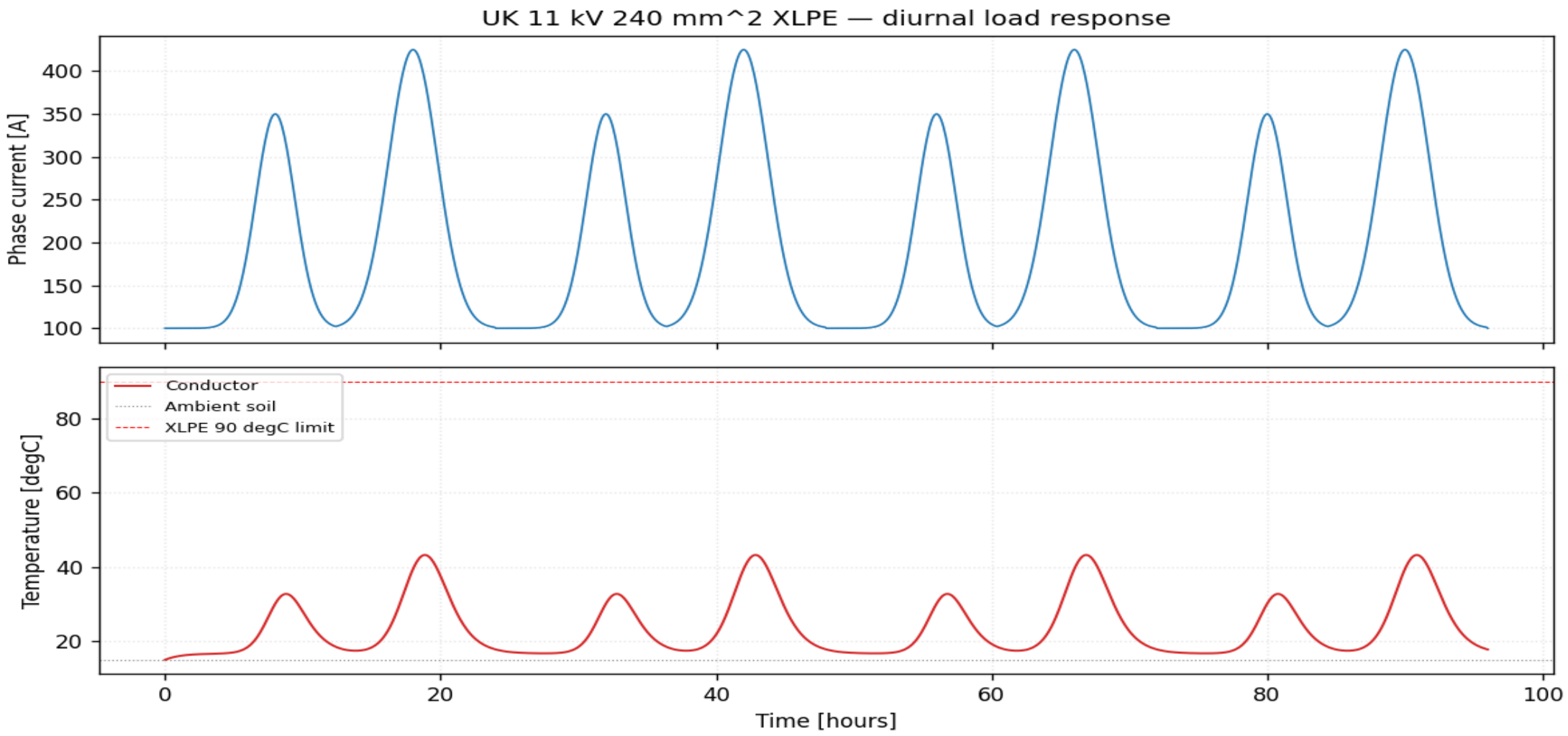

***Figure 5.*** *Four-day transient simulation under a UK two-peak residential load profile (100–350 A swing, Met Office winter ambient). Daily peak conductor temperature reaches 43 °C against the XLPE 90 °C thermal limit; the visible lag between load peaks and temperature peaks reflects the first-order thermal time constant of the cable approximately 64 minutes.*

Figure 5 shows a four-day simulated trace under a UK two-peak domestic load profile (residential profile, 100–350 A swing, Met Office winter ambient ~5 °C). The first-order time-constant of the cable, $\tau \approx 64$ min, is visible as the lag between load peaks and conductor-temperature peaks. Daily peak conductor temperature reaches 43 °C, well below the XLPE 90 °C thermal limit; mean conductor temperature 24 °C reflects typical UK distribution-feeder loading.

## 8.2 Long-horizon ageing

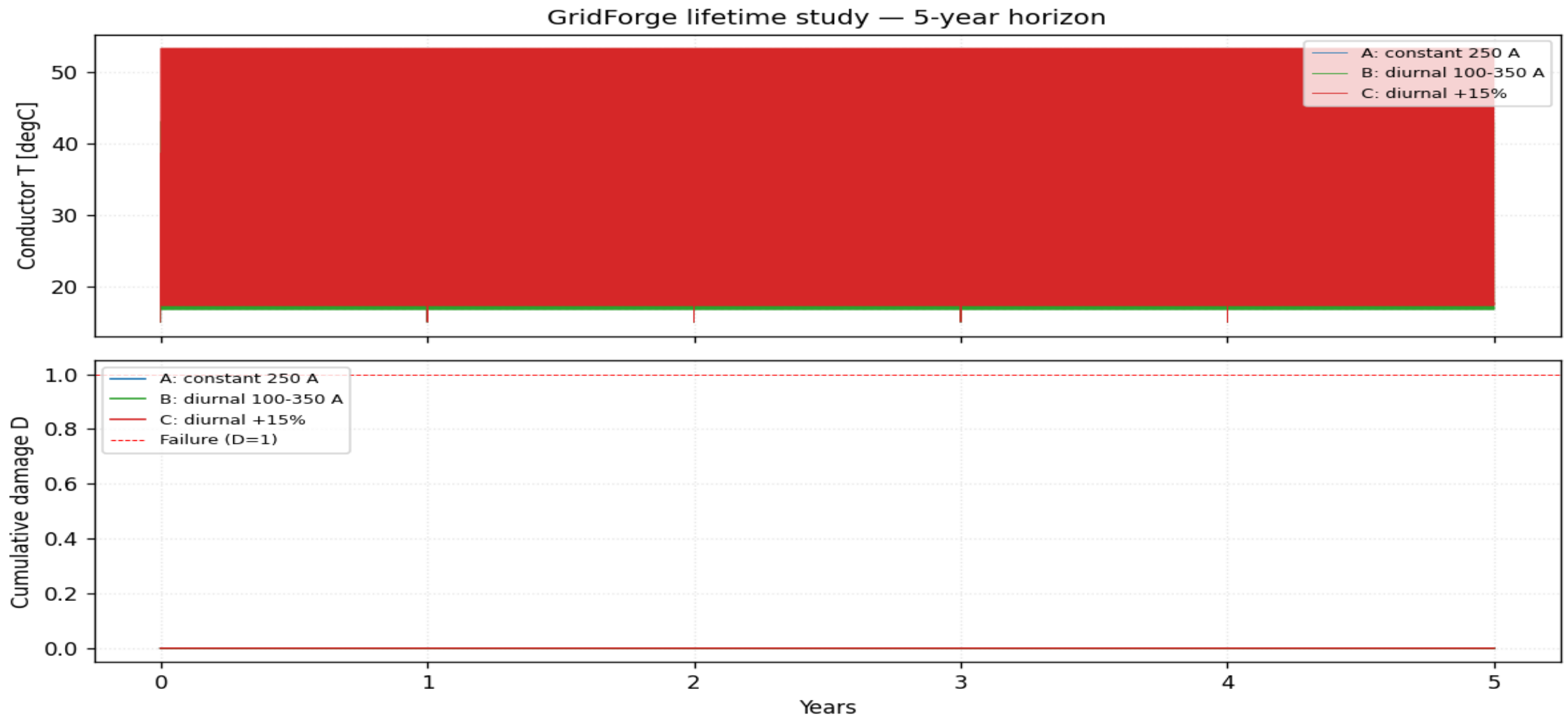

***Figure 6.*** *Cumulative Crine damage over five years for three operating scenarios: constant 250 A loading; the Section 8.1 diurnal profile; and the diurnal profile scaled by 1.15 to simulate moderate overload. Final damage at five years sits between $9 \times 10^{-8}$ and $3.6 \times 10^{-7}$, more than six orders of magnitude below the failure threshold of $D = 1$.*

Figure 6 plots cumulative Crine damage over 5 years for three operating scenarios: constant 250 A loading, the Section 8.1 diurnal profile, and the diurnal profile scaled by 1.15 to simulate moderate overload. Final damage at 5 years sits between $9 \times 10^{-8}$ and $3.6 \times 10^{-7}$, more than six orders of magnitude below the failure threshold of $D = 1$. At typical UK distribution loadings, **Crine-driven thermal ageing is not the limiting failure mode**; real distribution-cable failures are dominated by water-tree growth at joints, mechanical disturbance, and partial-discharge defects (modelled in GridForge as separate condition-mode injectors, Section 4).

### 8.3 Condition-mode separation

Across the 64-cable benchmark, the four condition-mode injectors produce cumulative damage trajectories that span four orders of magnitude at 5 years, ordered as expected: healthy < water_ingress < thermal_ageing < accelerated_dielectric. The same ordering holds within each archetype, with smaller-conductor cables (95 mm²) running hotter and ageing faster than larger ones (300 mm² single-core).

## 9 Limitations

GridForge v0.0.5 has four limitations that should be noted:

1. **Synthetic rather than real-world data.** All benchmark results are produced from a calibrated forward model. As such, the dataset does not capture phenomena commonly observed in DNO data, including joint defects, non-monotonic water-tree behaviour, measurement noise, and communication gaps. Addressing this limitation through real-data calibration, under partner DNO NDAs, is a priority future-work objective.
2. **Lumped transient treatment (T4).** The transient component is encapsulated in task T4, making it difficult for a purely steady-state surrogate to achieve zero T4 RMSE. A revised formulation incorporating both the transient dynamics and an explicit steady-state ground-truth column is planned.

3. **Single-cable thermal modelling.** The current framework does not include the IEC 60287-2-1 group-rating corrections for mutual heating effects (e.g. trefoil configurations), as these are not yet available. Consequently, the 300 mm² single-core archetype yields optimistic estimates of approximately 5–10 % under peak-load conditions.
4. **Simplified ageing physics for PILC.** The PILC archetype currently relies on Crine kinetics calibrated to XLPE systems. This does not reflect the distinct degradation mechanisms of paper insulation, whose ageing dynamics have yet to be incorporated.

### 10 Reproducibility

Every figure and table in this paper is reproducible from a clean checkout of the GridForge repository:

```
git clone https://github.com/mlily2024/Gridforge.git
cd Gridforge
pip install -e ".[dev,ml]"
pytest                              # 229 tests
python scripts/01_validate_iec60287.py
python scripts/02_transient_diurnal_load.py
python scripts/03_lifetime_estimate.py
python scripts/04_generate_mini_dataset.py
python scripts/05_train_pinn_iec_oracle.py
python scripts/06_run_benchmark.py
```

Random seeds are fixed throughout the codebase. The training seed of the PINN is 2026_04_28; the per-cable seed of the dataset is the cable index. Trained PINN weights are saved alongside the validation scatter at scripts/output/05_pinn_training/. The benchmark leaderboard CSV is written to scripts/output/06_benchmark/leaderboard.csv.

A pytest -m slow invocation reproduces the 0.5 °C acceptance test from cold weights and asserts the result.

## 10.1 Experimental platform

All experiments, including the roughly two-minute PINN training run, were carried out on a single machine: a 13th-generation Intel Core i5-1335U processor (10 cores, 12 threads), 8 GB of RAM, running Windows 11, with Python 3 and a CPU-only build of PyTorch, with no GPU acceleration. The IEC 60287 solver, the dataset generation, the PINN training, and the benchmark evaluation were all run on this platform. The training and run times reported in this paper were measured in this environment; run time is hardware-dependent and is not directly comparable across platforms, whereas the RMSE accuracy figures are platform-independent.

### 11 Conclusion

GridForge is a lightweight and purposefully constrained artefact designed to establish an open-benchmarking and open-physics framework within the UK distribution-cable monitoring community. Its long-term objective is to foster a culture of transparent, reproducible evaluation of decision support solutions for distribution-cables diagnostic monitoring and cable networks maintenance prioritisation. The framework is deliberately compact, enabling rapid adoption and low overhead, while remaining sufficiently expressive to support meaningful comparison of competing methods within a shared evaluation reference data space. At present the discriminating power of the benchmark is concentrated in the virtual-sensor task (T4): tasks T1, T3 and T5 tie because the reference baselines defer to shared downstream logic, and T2 is undefined on the current test split, so extending the baselines with task-specific logic is the immediate priority before the suite can rank models across all five tasks. Community engagement is integral to its evolution; thus critiques, counterarguments, and extensions are explicitly encouraged. For open validation, the dataset, test sets and models are made available at https://github.com/mlily2024/Gridforge.

## Acknowledgements

Public datasets and standards consulted include BS EN 60228, BS 7870, IEC 60287-1-1, IEC 60287-2-1, IEC 60853-2, IEEE Std 1407-2007, the UK Met Office 1991–2020 soil-temperature climatology, and the Open Networks Project secondary-substation study series.